\documentclass{article}

\usepackage{PRIMEarxiv}

\usepackage[utf8]{inputenc} 
\usepackage[T1]{fontenc}    
\usepackage{hyperref}       
\usepackage{url}            
\usepackage{booktabs}       
\usepackage{amsfonts}       
\usepackage{nicefrac}       
\usepackage{microtype}      
\usepackage{lipsum}
\usepackage{fancyhdr}       
\usepackage{graphicx}       
\graphicspath{{media/}}     
\usepackage{cancel}
\usepackage{acronym}
\usepackage{subcaption}
\usepackage{array}
\usepackage{tabularx}
\usepackage{booktabs}
\usepackage{multirow}
\usepackage{changepage}
\usepackage{float}

\acrodef{LfD}{Learning from Demonstration}
\acrodef{pHRI}{physical Human-Robot Interaction}
\acrodef{RL}{Reinforcement Learning}
\acrodef{IOC}{Inverse Optimal Control}
\acrodef{MP}{Movement Primitive}
\acrodef{DMP}{Dynamic Movement Primitive}
\acrodef{GP}{Gaussian Process}
\acrodef{GMM}{Gaussian Mixture Model}
\acrodef{ProMP}{Probabilistic Movement Primitive}
\acrodef{IRL}{Inverse Reinforcement Learning}
\acrodef{HMM}{Hidden Markov Model}
\acrodef{HRI}{Human-Robot Interaction}

\title{A Practical Roadmap to Learning from Demonstration for Robotic Manipulators in Manufacturing
}

\author{
  Alireza Barekatain $^{1}$, Hamed Habibi $^{1}$ and Holger Voos $^{1,2}$ \\
  $^{1}$ Automation and Robotics Research Group, Interdisciplinary Centre for Security, Reliability, and Trust (SnT),\\
  University of Luxembourg, Luxembourg \\
  $^{2}$ Faculty of Science, Technology, and Medicine, \\
  University of Luxembourg, Luxembourg \\
  \texttt{\{alireza.barekatain, hamed.habibi, holger.voos\}uni@lu} \\
}

\begin{document}

\maketitle

\begin{abstract}
This paper provides a structured and practical roadmap for practitioners to integrate \ac{LfD} into manufacturing tasks, with a specific focus on industrial manipulators. Motivated by the paradigm shift from mass production to mass customization, it is crucial to have an easy-to-follow roadmap for practitioners with moderate expertise, to transform existing robotic processes to customizable LfD-based solutions.
To realize this transformation, we devise the key questions of  "What to Demonstrate", "How to Demonstrate", "How to Learn", and "How to Refine".
To follow through these questions, our comprehensive guide offers a questionnaire-style approach, highlighting key steps from problem definition to solution refinement. The paper equips both researchers and industry professionals with actionable insights to deploy \ac{LfD}-based solutions effectively. By tailoring the refinement criteria to manufacturing settings, the paper addresses related challenges and strategies for enhancing \ac{LfD} performance in manufacturing contexts.
\end{abstract}


\section{Introduction}
In the context of robotics, Learning from Demonstration (LfD) refers to ``\textit{the paradigm in which robots acquire new skills by learning to imitate an expert}" \cite{ravichandar2020recent}, \textit{i.e.}, a robot learns to perform a task by watching a human's actions rather than being explicitly programmed.
\ac{LfD} allows robots to acquire new skills or refine existing ones while reducing the need for manual programming of robot behaviors, ultimately eliminating the requirement for a robotic expert \cite{ravichandar2020recent}.

\ac{LfD} offers a distinct approach to programming robots compared to traditional manual programming methods.
Traditional manual programming involves writing code or scripts to explicitly define the sequence of actions and movements for the robot to perform a task.
It requires expertise in robot programming languages, kinematics, and dynamics.
Moreover, writing code for complex tasks can be time-consuming, and any changes in the environment or task requirements demand extra re-programming effort \cite{heimann2020industrial}.
On another paradigm, optimization-based programming involves formulating the robot's task as an optimization problem, where the objective is to minimize or maximize a certain objective \cite{leger2016off}.
While this method optimizes the task execution based on various environments and task requirements and does not need re-programming, it still requires high robotic expertise to carefully formulate the task as an optimization problem and mathematically model the task and the environment in order to efficiently solve for the best solution.

\ac{LfD} surpasses these limitations by allowing non-experts to teach the robot without mathematical formulation or any robotic knowledge.
Also, the \ac{LfD} algorithm learns and generalizes the task flexibly according to the environment and task requirements.
Teaching a new task or refining a task takes significantly less effort, does not require robotic expertise, and can make the robotic system up and running in a relatively quicker manner.
These benefits have been found to be useful in the industry, where efficiency and agility gain importance when using robots for industrial tasks \cite{dean2016robotic, sanneman2021state}.

Several notable reviews and surveys focus on \ac{LfD} from different points of view and consider various aspects.
The work in \cite{ravichandar2020recent} gives a general review of \ac{LfD} and provides an updated taxonomy of the recent advances in the field. 
In \cite{fang2019survey} the authors survey LfD for robotic manipulation, discussing the main conceptual paradigms in the \ac{LfD} process.
In \cite{liu2022robot} the focus is on the applications of LfD in smart manufacturing.
They introduce and compare the leading technologies developed to enhance the learning algorithms and discuss them in various industrial application cases.
In \cite{zhu2018robot} the focus of the survey is on robotic assembly.
Specifically, they discuss various approaches to demonstrate assembly behaviors and how to extract features to enhance \ac{LfD} performance in assembly.
They also analyze and discuss various methods and metrics to evaluate \ac{LfD} performance.
Authors in \cite{sosa2022learning} survey \ac{LfD} advances with respect to human-robot collaborative tasks in manufacturing settings.
They provide detailed insights into the role of various human interactions in different \ac{LfD} methods.
The technical nature of the survey makes it suitable for active researchers in \ac{LfD} to seek new directions on making the \ac{LfD} algorithms more collaborative.

While the existing studies provide valuable insights into various aspects of \ac{LfD}, they often lack a comprehensive roadmap or practical guidance for practitioners. They mostly focus on presenting the state of the art without providing a clear roadmap for practitioners to implement \ac{LfD} in their robotic tasks.
Moreover, the technical depth of most of the studies requires a strong background in \ac{LfD}, making it inaccessible to non-academic practitioners or researchers who are new to the field.
Therefore, there is a need for a new study that not only consolidates existing knowledge but also bridges the gap between research and practice.

From a practical point of view, the recent advancements in the manufacturing industry create an increasing need for adaptable manufacturing robotic systems to perform flexibly with the variant demands of the market.
The production schemes are shifting from mass production, where a fixed line of manufacturing is used to create products on a mass scale, to mass customization, where production is in smaller batches of different products according to the market need  \cite{pedersen2016robot,cohen2019assembly}.
To retain the efficiency and cost-effectiveness of mass production schemes, robotic systems have to quickly adapt to new tasks and manufacturing requirements \cite{wind2001customerization,gavspar2020smart}. 
Such transition and requirements have led to a significant application of \ac{LfD}, as a suitable solution,  in the manufacturing industry.
It means that the existing robotic tasks in the mass production scheme need to be transformed via \ac{LfD} to meet the new requirements of mass customization, which is why it is necessary to provide guidance for industry practitioners to start deploying state-of-the-art \ac{LfD} solutions in existing robotic tasks.
Notably, among all the robotic systems, industrial manipulators are the most popular and versatile robot types widely used in manufacturing and production lines.
Therefore, while the application of robotic systems is not limited to manipulators, it is beneficial to have the focus of this paper narrowed down to industrial manipulators, due to their critical role in the automation of manufacturing and production lines.

Motivated by the aforementioned considerations and in contrast to existing reviews, our work aims to offer a practical and structured approach to implementing \ac{LfD} in manufacturing tasks. Unlike other reviews, our review takes the form of a comprehensive questionnaire-style guide, providing practitioners with a clear roadmap to integrate \ac{LfD}-based robot manipulation. Tailored for moderate expertise requirements, this tutorial-style taxonomy offers step-by-step instructions, enabling both researchers and professionals to develop application-based \ac{LfD} solutions. This review provides the readers with the main steps to define the problem and devise an \ac{LfD} solution, as well as giving main research directions for refining the performance of the \ac{LfD}.
The refinement criteria are also tailored based on the practical application of \ac{LfD}.

For this purpose, This paper explores how to integrate \ac{LfD} into the robotization process using manipulators for manufacturing tasks, depicted in Figure~\ref{fig:overall}. First, the practitioner addresses the question of ``What to demonstrate" to define the ``Scope of Demonstration". Subsequently, the practitioner needs to answer the question of ``How to Demonstrate" in order to devise a ``Demonstration Mechanism". Accordingly, the question of  ``How to learn" equips the robotic manipulator with the proper ``Learning mechanism". Even though the \ac{LfD} process implementation is accomplished at this stage, the evaluation of \ac{LfD} performance leads to the question of ``How to Refine", which provides the research objectives and directions in which the performance of the \ac{LfD} process can be further improved. Taking these points into account, the rest of the paper is structured as follows:

\begin{figure}[H]
\includegraphics[width=\linewidth]{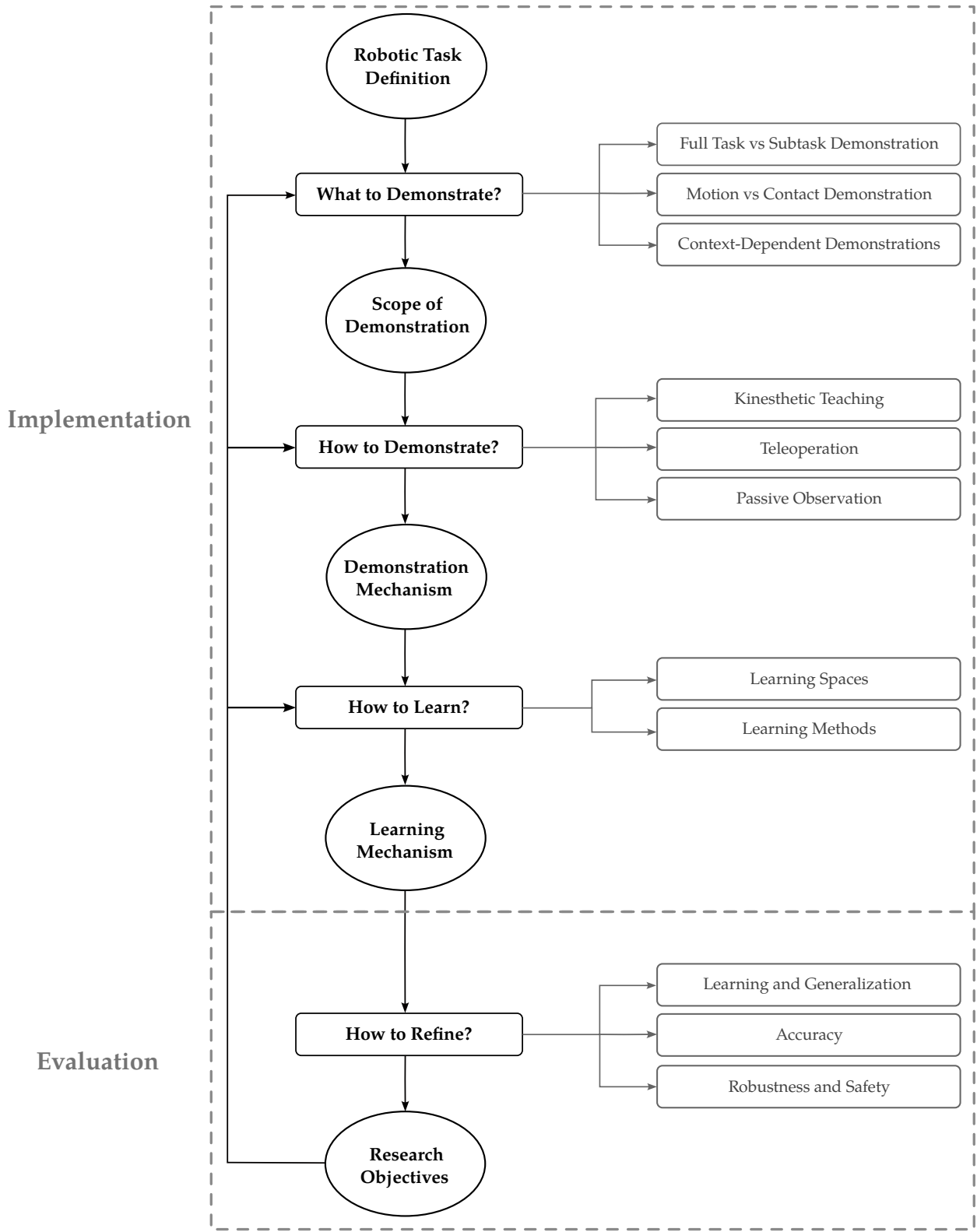}
\caption{Overview of our proposed roadmap for \ac{LfD} implementation.\label{fig:overall}}
\end{figure} 

\begin{enumerate}
    \item \textbf{What to Demonstrate?} (Section \ref{sec:WhattoDemonstrate}): This section explores how to carefully define the task and identify certain features and characteristics that influence the design of the process.
    \item \textbf{How to Demonstrate?} (Section \ref{sec:HowtoDemonstrate}): Building upon the scope of demonstration, this section explores effective demonstration methods, considering task characteristics and their impact on robot learning.
    \item \textbf{How to Learn?} (Section \ref{sec:HowToLearn}): In this section, the focus shifts to implementing learning methods, enabling autonomous task execution based on human demonstrations.
    \item \textbf{How to Refine?}  (Section \ref{sec:HowToRefine}): Concluding the structured approach, this section addresses refining \ac{LfD} processes to meet industrial manufacturing requirements, outlining challenges and strategies for enhancing \ac{LfD} solutions in manufacturing settings.
\end{enumerate}

\section{What to Demonstrate}
\label{sec:WhattoDemonstrate}
This section focuses on the first step in developing an \ac{LfD} solution, \textit{i.e.}, extracting the scope of the demonstration from the desired task. In this step, with a specific robotic task as the input, we explore how to determine the knowledge or skills a human teacher needs to demonstrate to the robot.
The scope of demonstration establishes clear boundaries for what the robot should be able to accomplish after learning.
Defining the scope is crucial because it sets the foundation for the entire \ac{LfD} process.
A well-defined scope ensures the provided demonstrations comprehensively and accurately capture the desired robot behavior.
Conversely, a poorly defined scope can lead to incomplete or inaccurate demonstrations, ultimately limiting the effectiveness of the \ac{LfD} solution.

To determine ``What to demonstrate", three different aspects of the robotic task will be investigated as follows.

\subsection{Full Task versus Subtask Demonstration}

\begin{figure}[H]
\includegraphics[width=\linewidth]{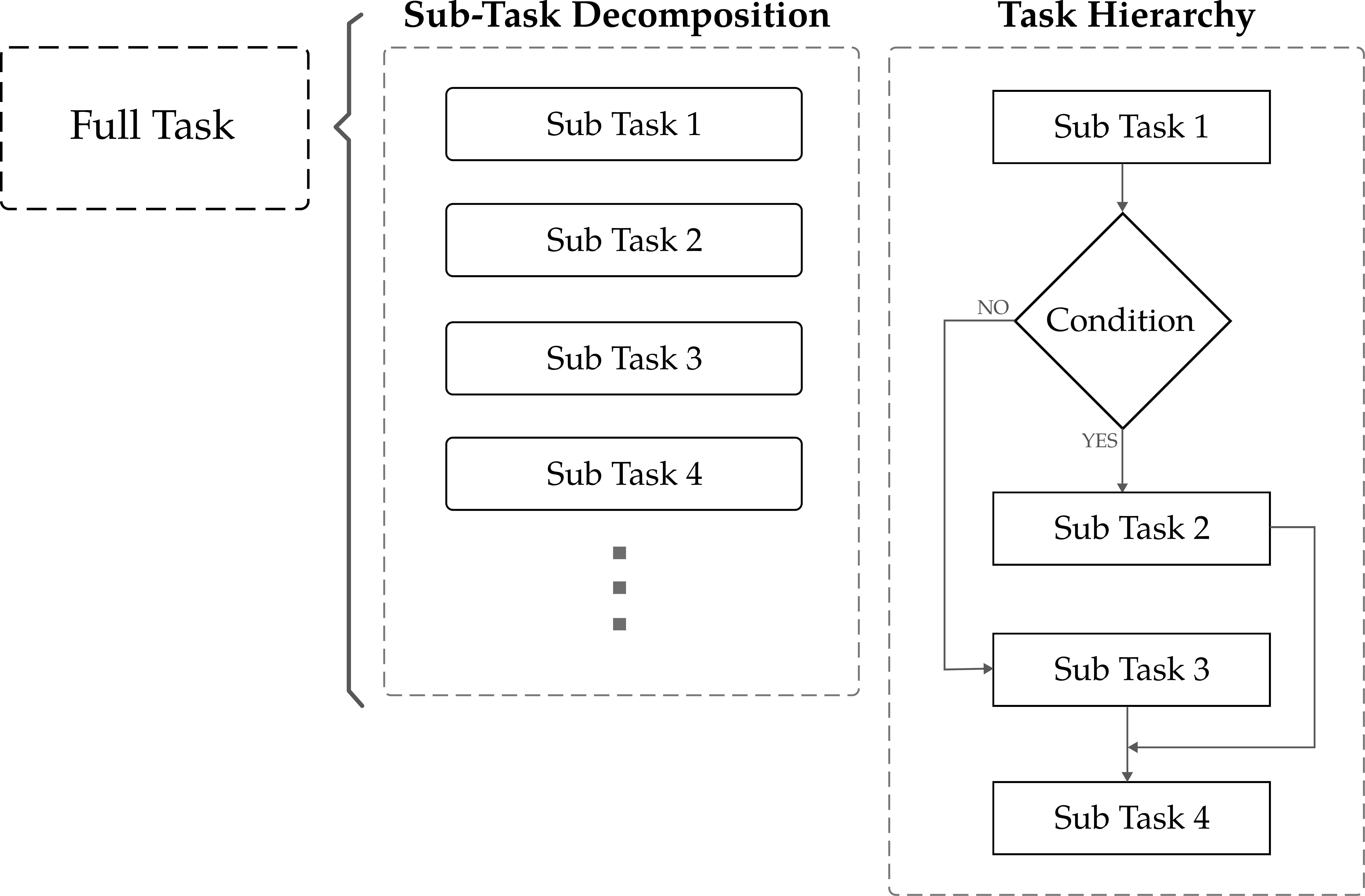}
\caption{Illustration of how subtasks and task hierarchy comprise a full task.\label{fig:task}}
\end{figure}

A full robotic task can be decomposed into smaller steps called subtasks, along with their associated task hierarchy or their logical sequence (Figure~\ref{fig:task}). In a full task demonstration, the human teacher demonstrates the entire process, including all subtasks in their logical order. In contrast, subtask demonstration focuses on teaching each subtask of the robotic task one at a time. Here we explore whether to demonstrate the full robotic task at once, or aim to demonstrate subtasks separately.

\textbf{What Happens When Learning Full Task:}
In the case of full task demonstration, the \ac{LfD} algorithm is required first to segment the task into smaller subtasks and learn them separately \cite{ekvall2008robot, origanti2021automatic, niekum2015learning, steinmetz2019intuitive, iovino2023framework, french2023super, willibald2022multi, mayershofer2023task, gugliermo2023learning, scherf2024learning, eiband2023online}.
Otherwise, training a single model on the entire task can lead to information loss and poor performance  \cite{lin2016movement}.
Beyond identifying subtasks, a full robotic task involves the logical order in which they should be executed – the task hierarchy. 
The \ac{LfD} algorithm is required to extract these relationships between subtasks to build the overall task logic \cite{ekvall2008robot, niekum2015learning, steinmetz2019intuitive, iovino2023framework, gugliermo2023learning, scherf2024learning}.
This segmentation is achieved through spatial and temporal reasoning on demonstration data  \cite{origanti2021automatic, steinmetz2019intuitive, french2023super, willibald2022multi, mayershofer2023task, gugliermo2023learning, eiband2023online, sorensen2023robot, dreher2022learning, zhou2012hierarchical,xiong2016robot,carpio2019learning}.
Spatial features help identify subtasks, while temporal features reveal the high-level structure and sequence.

For instance, consider demonstrating a pick-and-place task as a full task.  The \ac{LfD} algorithm can easily segment the demonstration into ``reaching", ``moving", and ``placing" the object, along with their sequential task hierarchy. Because these subtasks involve clearly defined motions and follow a straightforward, linear order, the \ac{LfD} algorithm can reliably extract the complete logic required to perform the entire pick-and-place task.
On the other hand, consider a pick-and-insert task with tight tolerances. Precise insertion is challenging to demonstrate and requires retry attempts as recovery behavior. This creates a conditional task hierarchy.
The successful insertion depends on achieving tight tolerances, and if the initial attempt fails, the \ac{LfD} system needs to learn the recovery behavior of repositioning the object and attempting insertion again.
Consequently, \ac{LfD}'s reliance on automatic segmentation to extract the detailed task logic in such cases becomes less reliable.
However, background information on the task characteristic can be provided as metadata by the human teacher and leveraged by the learning algorithm to improve the segmentation method in semantic terms  \cite{gustavsson2022combining, willibald2022multi}.

\textbf{What Happens When Learning Subtask:}
When subtasks are demonstrated individually, the human teacher manually breaks down the full task and provides separate demonstrations for each one, along with the associated task hierarchy. This isolates the \ac{LfD} algorithm's learning process, allowing it to focus on learning one specific subtask at a time \cite{ewerton2016incremental, pignat2019bayesian, maeda2017active, wang2024robot, meszaros2022learning, koert2019learning, raiola2015co}. It is evident that this approach requires extra effort from the teacher compared to full task demonstration.

For complex or intricate tasks such as pick-and-insert with tight tolerances, the teacher can individually provide ``reaching", ``moving", and ``inserting" subtasks along with recovery behaviors. While this requires the teacher to separately define the task hierarchy and provide demonstrations for each subtask, it yields several benefits. First, the teacher can focus on providing clear and accurate demonstrations for each isolated step. Second, it allows for a reliable demonstration of the conditional hierarchy involved in complex tasks  \cite{mohseni2019simultaneous,bobu2024aligning}.

\textbf{When to Demonstrate Full Task versus Subtask:}
Demonstrating the entire robotic task at once can be efficient for teaching simple, sequential tasks, as the algorithm can segment and learn reliably. However, for complex tasks with conditional logic or unstructured environments, this approach struggles. Breaking the task into subtasks and demonstrating them individually is more effective in these cases. This removes the burden of segmentation from the learning algorithm and allows for better performance, especially when dealing with conditional situations. By teaching subtasks first, and then layering the task hierarchy on top, robots can handle more complex tasks and learn them more efficiently. In essence, full task demonstration is recommended for simple behaviors with linear task logic, while complex tasks are recommended to be decomposed into subtasks and demonstrated separately.

\subsection{Motion-Based versus Contact-Based Demonstration}
\begin{figure}[H]
\includegraphics[width=\linewidth]{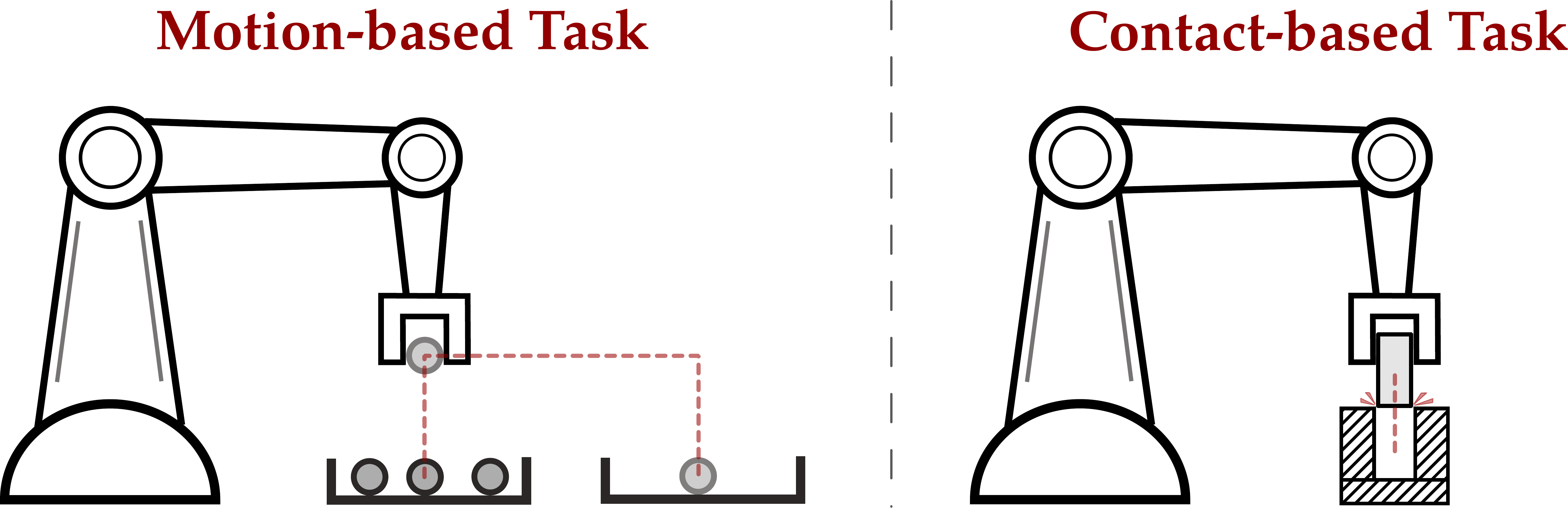}
\caption{Comparison of motion-based versus contact-based task. On the left, the pick-and-place task has a structured and predictable interaction with the environment, while the insertion task on the right needs to deal with the contact resulting from tight tolerances to successfully perform the task. \label{fig:motion_force}}
\end{figure}

Robot task demonstrations can be categorized into two main types: motion-based and contact-based. Motion-based tasks, like pick-and-place  \cite{meszaros2022learning, koert2019learning, raiola2015co}, focus on teaching the robot's movement patterns. The key information for success is captured in the robot's trajectory and kinematics, with limited and controlled contact with the environment \cite{meszaros2022learning,koert2019learning,raiola2015co,dong2022passive,liu2021robotic,mo2023multi,frank2021constrained,zhai2022motion,auddy2023continual,ruan2024primp,biagiotti2023robot}. 
Conversely, contact-based tasks, such as insertion   \cite{vuong2021learning, wu2023prim, johannsmeier2019framework, simonivc2021analysis, wu2021learning, lee2019making,davchev2022residual, vecerik2017leveraging, pignat2019bayesian, vidakovic2020accelerating, perico2019combining, roveda2020assembly, si2022adaptive}, require the robot to understand how to interact with objects. Here, task success also relies on understanding forces and contact points  \cite{si2022adaptive}. Simply replicating the motion does not suffice and the robot needs to learn to apply appropriate force or adapt to tight tolerances to avoid failure. This highlights the importance of contact-based demonstrations for tasks where interaction with the environment is crucial. The comparison of motion-based and contact-based tasks is illustrated in Figure~\ref{fig:motion_force}.

\begin{figure}[H]
\includegraphics[width=\linewidth]{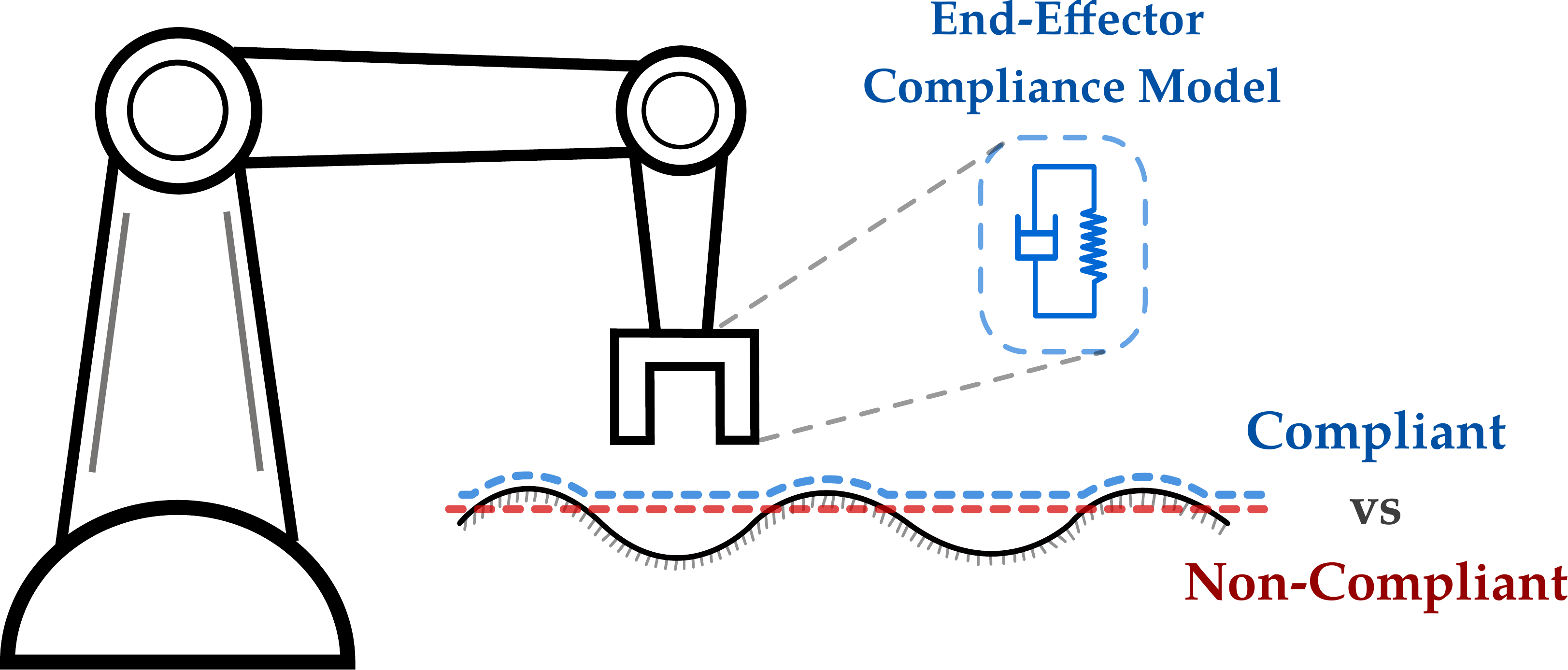}
\caption{Illustration of compliant versus non-compliant behavior against the environment. The black line represents the environment surface, the red path represents a non-compliant behavior, and the blue path represents the compliant behavior against the environment surface. In Impedance Control, the end-effector is modeled as a spring-damper system.\label{fig:compliance}}
\end{figure}

\textbf{Notion of Compliance:}
To understand whether a task is motion-based or contact-based, it is necessary to understand the notion of compliance.
Compliance in robotics refers to the capacity of a robotic system to yield or adjust its movements in response to external forces, ensuring a more adaptable and versatile interaction with its environment \cite{wang1998passive}.
This adaptability is typically achieved via Impedance Controllers, where the end effector of the robot is modeled as a spring-damper system to represent compliance (Figure~\ref{fig:compliance})  \cite{song2019tutorial,hogan1984impedance}.
In motion-based tasks, the robot prioritizes following a planned path with minimal adjustment (low compliance), \textit{i.e.} external forces cannot alter the robot's behavior, while contact-based tasks allow for more adaptation (high compliance) to better interact with the environment. It is important not to confuse compliance with collision avoidance. Collision avoidance involves actively preventing contact with the environment by adapting the behavior on the kinematic level, while compliance relates to the robot's ability to adjust its behavior in response to external forces.

\textbf{What is Learned and When to Teach:}
With motion-based demonstration, the \ac{LfD} algorithm learns under the assumption of zero compliance and replicates the behavior on a kinematic level, \textit{i.e.}, strict motion.
On the other hand, contact-based teaching enables the \ac{LfD} algorithm to learn how to react when compliance is high, therefore it learns the skills on how to interact with the working environment.

One critical factor in selecting between motion-based and contact-based demonstration is the analysis of the desired task through the lens of its dependence on compliance.
For example, in an insertion task with tight tolerances, if there are slight inaccuracies in measurements and the peg lands with a slight offset to the hole, compliance can be helpful to react skillfully to this misalignment and attempt to find the right insertion direction.
Conversely, a pick-and-place task typically does not require compliance in the case when the grasping mechanism is simple and structured.

\textbf{Practical Implications:}
From the point of view of practical implementation, the trade-off between motion and contact during the task execution is a key design factor to implement the task successfully \cite{origanti2021automatic,pignat2019bayesian,si2022adaptive,roveda2020assembly,nemec2018efficient,hu2022robot,seo2023contact}.
For example, in tasks such as rolling dough  \cite{si2022adaptive}, board wiping  \cite{pignat2019bayesian}, and grinding  \cite{origanti2021automatic}, hybrid motion and force profile are learned as both are crucial for successful task execution \textit{i.e.} a force profile is needed to be tracked alongside the motion.
As mentioned, this factor is best encoded via Impedance Control, where at each point of task execution, it can be determined whether position or force requirements are in priority \cite{wu2023prim,kastritsi2018progressive}. However, this method requires torque-controlled compliant robots and cannot be applied to many industrial robots which are position-controlled and stiff \cite{simonivc2021analysis}.
For such robots, the alternative to Impedance Control is Admittance Control, which operates with an external force sensor and can be implemented on stiff industrial robots \cite{yang2023learning}.

\subsection{Context-Dependent Demonstrations}
In addition to the choice of scope between full task versus subtask and motion versus contact demonstration, the operation of the robot is influenced by various specific contexts. Such contextual settings are highly dependent on the requirements of the task and often need to be custom-designed and tailored for the task. Nonetheless, here we discuss several common contexts alongside the considerations required for each.

\subsubsection{Collaborative Tasks}
Tasks that involve collaborating with humans or other robots typically provide interaction through an observation or interaction interface, which serves as a channel for information exchange between the robot and its collaborators \cite{wang2018facilitating,nemec2018human,koert2019learning,eiband2023collaborative,rozo2016learning,jha2023generalizable}. These interfaces can take various forms, such as physical interaction mechanisms or dedicated communication protocols  \cite{rozo2016learning, khoramshahi2019dynamical, Jahanmahin2022human, xing2023dynamic}, and are specifically designed and tailored to facilitate the collaborative task.
Consequently, when providing demonstrations for such tasks, careful consideration must be given to ensure alignment with the interaction interface.

A typical example of a collaborative task is collaborative object transportation where one side of the object is held by the robot and the other part is held by the human  \cite{nemec2018human}. In this case, the interaction interface is \ac{pHRI}, where not only the motion is important, but also the compliance becomes relevant. Another aspect to consider in collaborative tasks is safety and collision avoidance since the robot's operating environment is closely shared with the human collaborator  \cite{koert2019learning}. It means that the human teacher needs to further teach safety strategies to the robot. Moreover, collaborative tasks have a more complicated task logic since the execution can depend on the collaborator's actions, which adds more conditions and more branching in the logic of the task. It also requires the robot to predict the intention of the human in order to follow the task hierarchy  \cite{khoramshahi2019dynamical}. 

\subsubsection{Bi-Manual Tasks}

Bi-manual tasks involve the coordinated use of both robot arms to manipulate objects or perform activities that require dual-handed dexterity \cite{nemec2018efficient,dong2022passive,franzese2023interactive,krebs2021kit,stepputtis2022system,dreher2022learning}.
Teaching a robot to perform bi-manual tasks through \ac{LfD} should emphasize synchronization and coordination between the robot's multiple arms. For instance, in tasks like assembling components or handling complex objects, the robot needs to learn how to distribute the workload efficiently between its arms. Also, in dual-arm assembly, the coordination of both arms played a crucial role in achieving the desired precision \cite{nemec2018efficient,liu2023birp,krebs2021kit}

Moreover, Bi-manual tasks often require specialized grasping strategies if both arms are simultaneously used for manipulating items \cite{mao2023learning,stepputtis2022system}.
Given the proximity of both arms in bi-manual tasks, safety considerations become of great importance.
The teaching should emphasize safe practices, including collision avoidance strategies and safety-aware learning.

\subsubsection{Via points}
In certain contexts, teaching robot-specific via-points within a task can be a highly effective way to convey nuanced information and refine the robot's execution \cite{vidakovic2020accelerating,jaquier2020learning,arduengo2023gaussian}.
Via points serve as intermediate locations or configurations within a task trajectory, guiding the robot through critical phases or ensuring precise execution.
One notable scenario where demonstrating via points can enhance the learning process is in assembly processes \cite{ding2020task}.
For complex machinery assembly, instructors can guide the robot through specific via points to ensure proper component alignment or correct part insertion.
Additionally, via-points serve as an excellent measure of accuracy to optimize the robot motion and tool manipulation while ensuring successful task execution as long as the via point is passed throughout the path.
Those enable the learning algorithm to understand optimal trajectories, adapt to changing conditions, and enhance its overall versatility. 

\subsubsection{Task Parameters}
Some contexts require explicitly teaching a robot specific task parameters to enhance its understanding and performance in specialized scenarios.
These task parameters go beyond the general actions and involve teaching the robot how to adapt to specific conditions or requirements.
Here, the focus is on tailoring the robot's learning to handle variations in the environment, object properties, or operational constraints \cite{rozo2016learning,kulak2021active,prados2024learning,zappa2023parameterization,arduengo2023gaussian}.

Teaching the robot about variations in object properties is essential for tasks where the characteristics of objects significantly impact the manipulation process.
For instance, in material handling tasks, the robot needs to learn how to handle objects of different shapes, sizes, weights, and materials  \cite{cui2022coupled, kulak2021active}.
Demonstrations can be designed to showcase the manipulation of diverse objects, allowing the robot to generalize its learning across a range of scenarios.
Additionally, robots operating in manufacturing settings often encounter specific operational constraints that influence task execution.
Teaching the robot about these constraints ensures that it can adapt its actions accordingly.
Examples of operational constraints include limited workspace, restricted joint movements, or specific safety protocols \cite{li2022learning}.

\section{How to Demonstrate}
\label{sec:HowtoDemonstrate}

The next step after answering the question of ``What to Demonstrate" and defining the scope of demonstration is to realize how to provide demonstrations to transfer the required knowledge and skills from the human teacher to the robot, \textit{i.e.}, the channel through which the information intended by the teacher could be efficiently mapped to the \ac{LfD} algorithm on the robot.
According to  \cite{ravichandar2020recent}, demonstration methods can be classified into three main categories: Kinesthetic Demonstration, Teleoperation, and Passive Observation.
In this section, we discuss these categories and analyze their advantages and disadvantages with respect to the scope of demonstration, with a summary provided in Table~\ref{tab:how_to_demo}.

\subsection{Kinesthetic Teaching}

Kinesthetic teaching is a method wherein a human guides a robot through a desired motion within the robot's configuration space.
In this approach, a human physically guides the robot to perform a task, and the robot records the demonstration using its joint sensors (Figure~\ref{fig:kinesthetic}) \cite{ewerton2016incremental, meszaros2022learning, koert2019learning, wu2023prim,simonivc2021analysis, vecerik2017leveraging, origanti2021automatic, johns2021coarse, eiband2023collaborative, perico2019combining, shi2021combining, wohlgemuth2024electromyography, prados2023kinesthetic}.
The key aspect is the direct interaction between the human teacher and the robot in the robot's configuration space.

\textbf{What Setup is Required:}
Kinesthetic teaching offers a straightforward setup, requiring only the robot itself. This simplicity contributes to ease of implementation and reduces the complexity of the teaching process, as well as minimizing the associated costs. This makes kinesthetic teaching an affordable and cost-effective option for training robots. Moreover, the interaction with the robot is intuitive for the teacher, making it easier to convey complex tasks and subtle details.

However, the suitability of kinesthetic teaching can be limited by the physical demands it requires, particularly with larger or heavier robots. Safety concerns also arise, especially in scenarios involving rapid movements or the handling of hazardous materials. This limitation can affect the scalability of kinesthetic demonstrations in diverse manufacturing contexts.

\textbf{How Demonstration Data is Obtained:}
Through kinesthetic teaching, the robot records the demonstration using its joint sensors.
The recorded data forms the basis for training the robot, allowing it to learn and replicate the demonstrated motion. 
The mapping of training data into the learning algorithm is straightforward, which enhances the reliability of the demonstration framework.
However, the recorded demonstration data contain noise, as it depends on the physical interaction between the human and the robot.
This noise can affect the smoothness of the training data and require additional processing to improve the learning algorithm's performance \cite{barekatain2023dfl}.
Additionally, since the training data depends on the robot hardware, the scalability of the training data to another setup will be limited.

\textbf{Recommendations:}
Kinesthetic teaching is effective for instructing both full task hierarchies and low-level subtasks, especially excelling in demonstrating complex and detailed subtasks with its precise physical guidance of the robot. However, for full task demonstrations, additional post-processing of training data is advised to enhance segmentation and eliminate noise. While kinesthetic teaching offers precise control for motion demonstrations, the recorded data often suffers from noise and lacks smoothness due to the physical interaction between human and robot. However, it becomes limiting for contact-based demonstrations because unreliable torque readings from joint sensors prevent teaching the desired force profile to the robot, as the human guides its movements.


\begin{table}[H]
\caption{Summary of the comparison of demonstration mechanisms.\label{tab:how_to_demo}}
	\begin{adjustwidth}{-2cm}{0cm}
		\newcolumntype{C}{>{\centering\arraybackslash}X}
		\begin{tabularx}{\linewidth}{CCCC}
			\toprule
			\textbf{}	& \textbf{Kinesthetic Teaching}	& \textbf{Teleoperation}     & \textbf{Passive Observation}\\
                \midrule
                Concept & Physically guiding robot & Remotely guiding robot & Observing human actions\\
			\midrule
\multirow[m]{4}{*}{Advantages}	& Demonstrate Complex Motion	& Safe Demonstration			& Safe Demonstration\\
			  	          & Minimal Setup			      & Isolation of Teaching		  & Ease of Demonstration\\
			             	& Intuitive Interaction			& 			                    & \\
			             	& Precise Manipulator Control	& 			                    & \\
                   \midrule
\multirow[m]{2}{*}{Limitations}     & Safety Concerns			& Complex Setup			& Complex Setup\\
			  	               & Physically Demanding	 & Requires Skills to Use		& Inefficient for Complex tasks\\
                   \midrule
\multirow[m]{3}{*}{Recommended Use}   & Full Task Demonstration	   & Contact-Based Demonstration			& Full Task Demonstration\\
			  	               & Subtask Demonstration			& Iterative Refinement			&  Large-Scale Data Collection\\
			             	   & Motion Demonstration			& 			                    & \\
			\bottomrule
		\end{tabularx}
	\end{adjustwidth}
\end{table}

\subsection{Teleoperation}

Teleoperation refers to the process in which the human teacher remotely controls the movements and actions of the robot (Figure~\ref{fig:teleoperation}). This control can be facilitated through different methods including joysticks, haptic interfaces, or other input devices, enabling the operator to teach the robot from a distance \cite{pignat2019bayesian, si2022adaptive, si2021review, rigter2020framework, tung2021learning, luo2023vision, guleccyuz2023netlfd}. This method differs from kinesthetic teaching as it allows for remote teaching to the robot.

\textbf{What Setup is Required:}
Setting up teleoperation involves equipping the robotic manipulator with necessary sensors like cameras and force/torque sensors to relay feedback about the robot and its surroundings. On the human side, a well-designed control interface is required for intuitive and accurate control of the robot's movements. A robust communication system, whether wired or wireless, is necessary for real-time transmission of control signals and feedback. Safety measures, including emergency stop mechanisms, are essential to prevent unexpected behaviors, especially since the robot is operated remotely and immediate access to the robot is not possible in case of a malfunction.

The teleoperation setup is inherently well-suited for the tasks being operated in dangerous or hard-to-reach environments.
Moreover, the design of the teleoperation interface can be versatile according to the target task, to provide the operator with a teaching interface closest to human-like dexterity.
On the downside, teleoperation requires a more sophisticated setup compared to kinesthetic teaching, and it requires further designing the control and the communication interface according to the task.
While it can promote intuitive teaching, it often requires extra training for the operator on how to use the setup for their teaching and demonstration.
The remote nature of teleoperation can also raise concerns over the communication latency of the setup and how it affects the task demonstration depending on the task. 

\textbf{How Demonstration Data is Obtained:}
Through teleoperation, the acquisition of the training data can be flexibly designed based on what information is required for learning.
The training data can be obtained by joint sensor readings, force/torque sensors on the joints or the end effector, haptic feedback, etc.
It is the decision of the robotic expert how to map the raw teaching data to processed and annotated training data suitable for the \ac{LfD} algorithm.
One main advantage of teleoperation is that it is possible to isolate the teaching to a certain aspect of the task.
For example, in  \cite{meszaros2022learning}, a joystick is used as the teleoperation device, where certain buttons only adjust the velocity of the robot, and other buttons directly affect the end-effector position and leave the execution velocity untouched.
This benefit can enable better-tailored teaching or iterative feedback to increase the efficiency of the learning process.

\textbf{Recommendations:}
Teleoperation is a suitable approach for demonstrating contact-rich tasks, since there is no physical interaction, and the joint torque sensors or the end effector force sensor on the robot can reliably record the contact-based demonstration as training data. It is also well-suited for tasks demanding real-time adjustments.
One of the most common combinations of demonstration approaches is to use kinesthetic demonstration for motion demonstration, and use teleoperation for iterative refinements or providing contact-based demonstrations  \cite{franzese2021ilosa}.

\subsection{Passive Observation}

\begin{figure}[H]
    \centering
    \begin{subfigure}[b]{0.32\linewidth}
        \includegraphics[width=\linewidth]{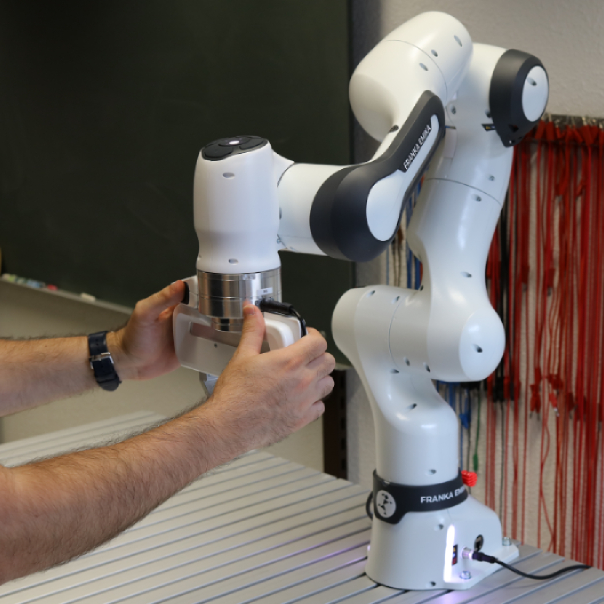}
        \caption{Kinesthetic Teaching}
        \label{fig:kinesthetic}
    \end{subfigure}
    \begin{subfigure}[b]{0.32\linewidth}
        \includegraphics[width=\linewidth]{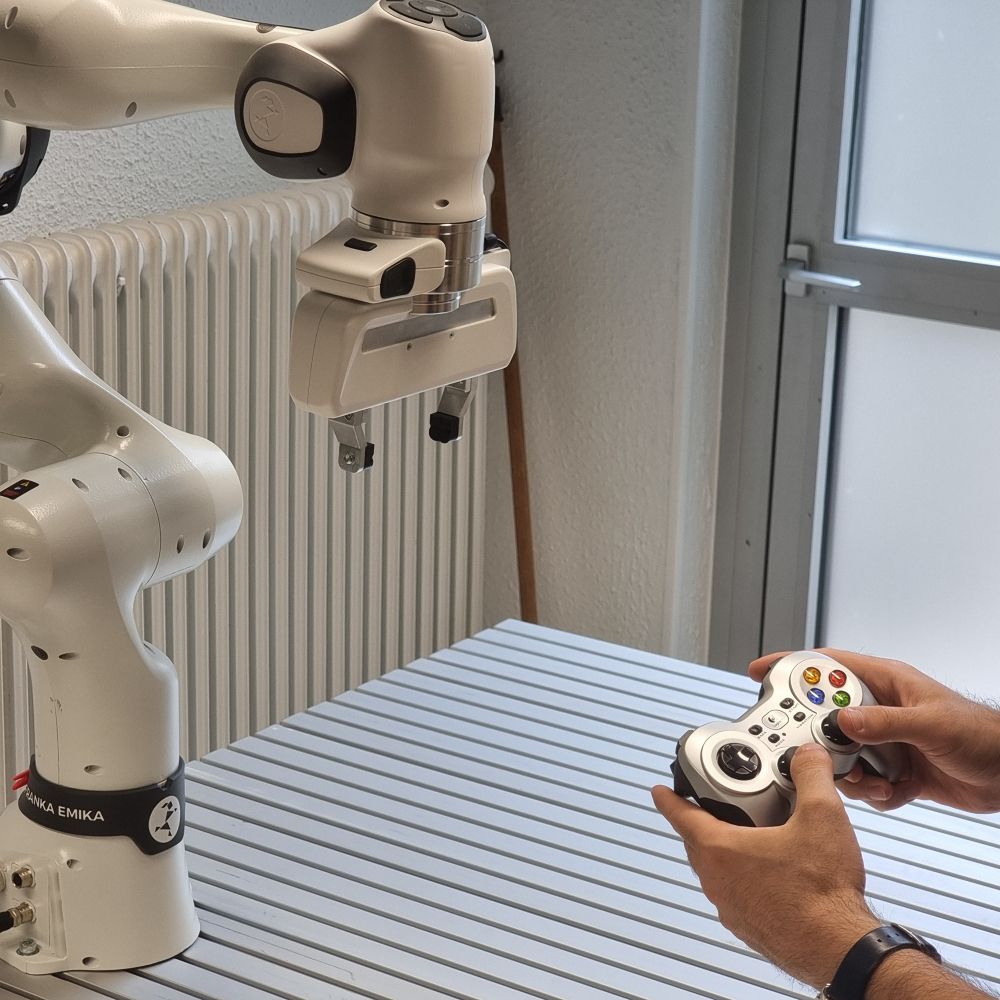}
        \caption{Teleoperation}
        \label{fig:teleoperation}
    \end{subfigure}
    \begin{subfigure}[b]{0.32\linewidth}
        \includegraphics[width=\linewidth]{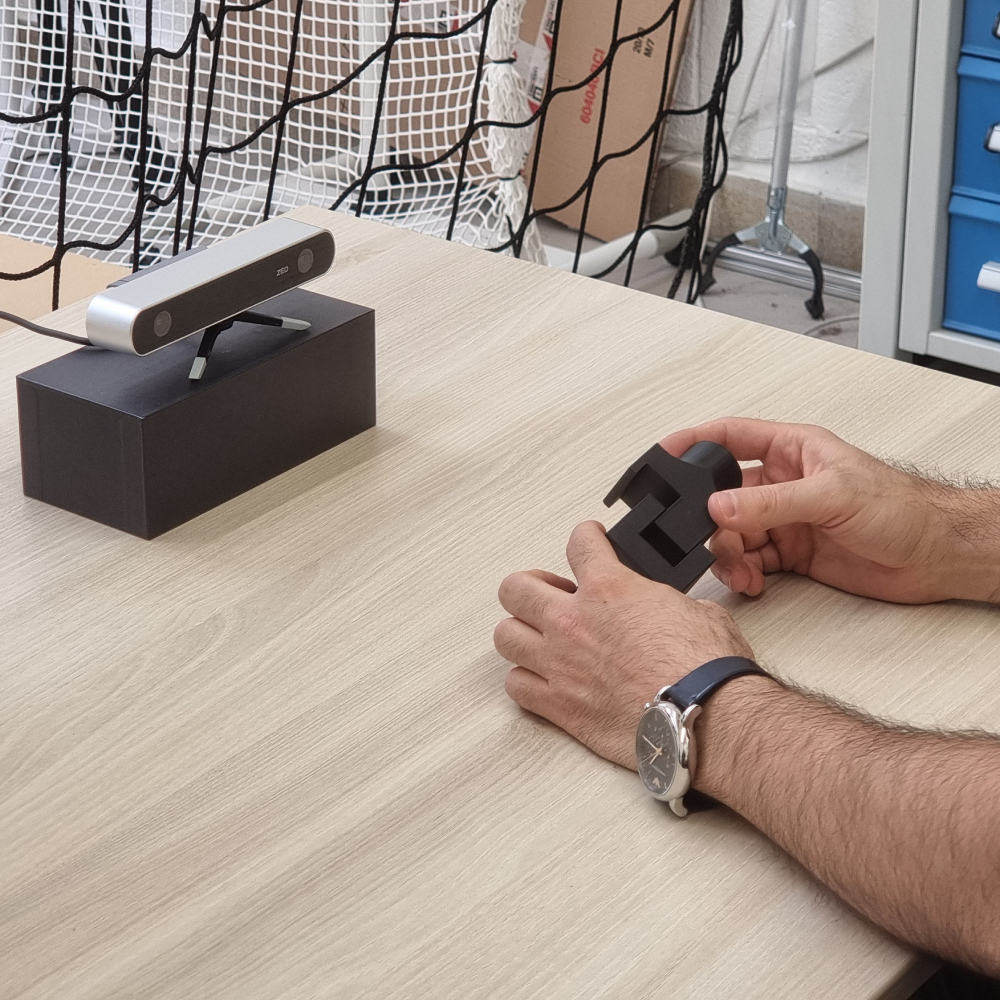}
        \caption{Passive Observation}
        \label{fig:passive}
    \end{subfigure}
    \vspace{1mm}
    \caption{Illustrative examples of the main demonstration approaches.}
    \label{fig:demo}
\end{figure}

Passive observation refers to the process of a robot learning by observing and analyzing the actions performed by a human or another source without direct interaction or explicit guidance, \textit{i.e.} the teaching happens with the robot outside the loop while passively observing via various sensors (Figure~\ref{fig:passive}) \cite{nemec2018efficient,yin2023multi, mayershofer2023task, zhu2023diff, yang2023watch, xu2022robot, yin2023multi, krebs2021kit}.
During passive observation, the robot captures and analyzes the relevant data, such as the movements, sequences, and patterns involved in a particular task.
The features and characteristics of the task are extracted from the observation and fed into the \ac{LfD} algorithm as training data for learning and generalization.

\textbf{What Setup is Required:}
Setting up the teaching framework for passive observation involves various sensors such as 2D/3D cameras, motion capture systems, etc. to enable the \ac{LfD} algorithm to observe the environment and the actions performed by the human teacher. 
The information from raw observation is then processed by sophisticated machine learning and computer vision algorithms to extract key features of the demonstration and track them throughout the teaching.
In this setup, humans often teach the task in their own configuration space (\textit{i.e.} with their own hands and arms), which makes the teaching easy and highly intuitive for the humans.
However, the setup is complicated and expensive due to the requirement of various sensory systems.

When it comes to the scalability of demonstration across numerous tasks and transferability from one robotic platform to another, passive observation stands out as a suitable option for large-scale collections of demonstration datasets for various tasks, making it preferable when extensive datasets are needed for training purposes.
This is while kinesthetic teaching and teleoperation mainly rely on a specific robotic platform and often cannot be scaled across tasks or robots.

\textbf{How Demonstration Data is Obtained:}
Acquisition of training data through passive observation mainly relies on the extraction of the key features, as the learning performance critically depends on how well the features represent the desired behavior of the robot on the task.
This forms a bottleneck in learning, since the more complex the task, the less efficient the feature extraction.
Consequently, passive observation can suffer from learning and performance issues when it comes to complex and detailed demonstrations.
Overall, as the demonstration setup is complex for this approach and the correspondence problem limits the possibility of demonstrating complex tasks, this approach is not common for manufacturing use cases.

\textbf{Recommendations}
Nonetheless, Passive observation has been used for demonstrating high-level full-task hierarchies. It is a suitable approach in scenarios where a diverse range of demonstrations across various tasks needs to be captured. For instance,  in  \cite{nemec2018efficient}, human demonstrations are observed via a Kinect motion tracker, and then a motion segmentation is applied to build the overall task logic. In terms of scalability, passive observation stands out as a fit option for large-scale data collection, making it particularly suitable when extensive datasets are needed for training purposes.

\subsection{Remarks}
There are alternative approaches to provide demonstrations tailored to a specific context or situation.
For example, in  \cite{biyik2023active} the demonstration is in the form of comparison between trajectories, \textit{i.e.} the robot produces a set of candidate trajectories, while the human provides preferences and comparison among them to imply which robot behavior is more suitable.
In  \cite{celemin2019interactive}, the human feedback has the form of a corrective binary signal in the action domain of the \ac{LfD} algorithm during robot execution.
The binary feedback signifies an increase or decrease in the current action magnitude.

\section{How to Learn}
\label{sec:HowToLearn}
This section focuses on the development of the \ac{LfD} algorithm itself, after determining the scope of demonstration and the demonstration mechanism. The aim of this section is to consider how to design and develop a learning mechanism to meet the requirements of our desired task. We first discuss the possible learning spaces in which the robot can learn, and then we explore the most common learning methods used as the core of \ac{LfD} algorithm.

\subsection{Learning Spaces}
\begin{figure}[H]
\includegraphics[width=0.85\linewidth]{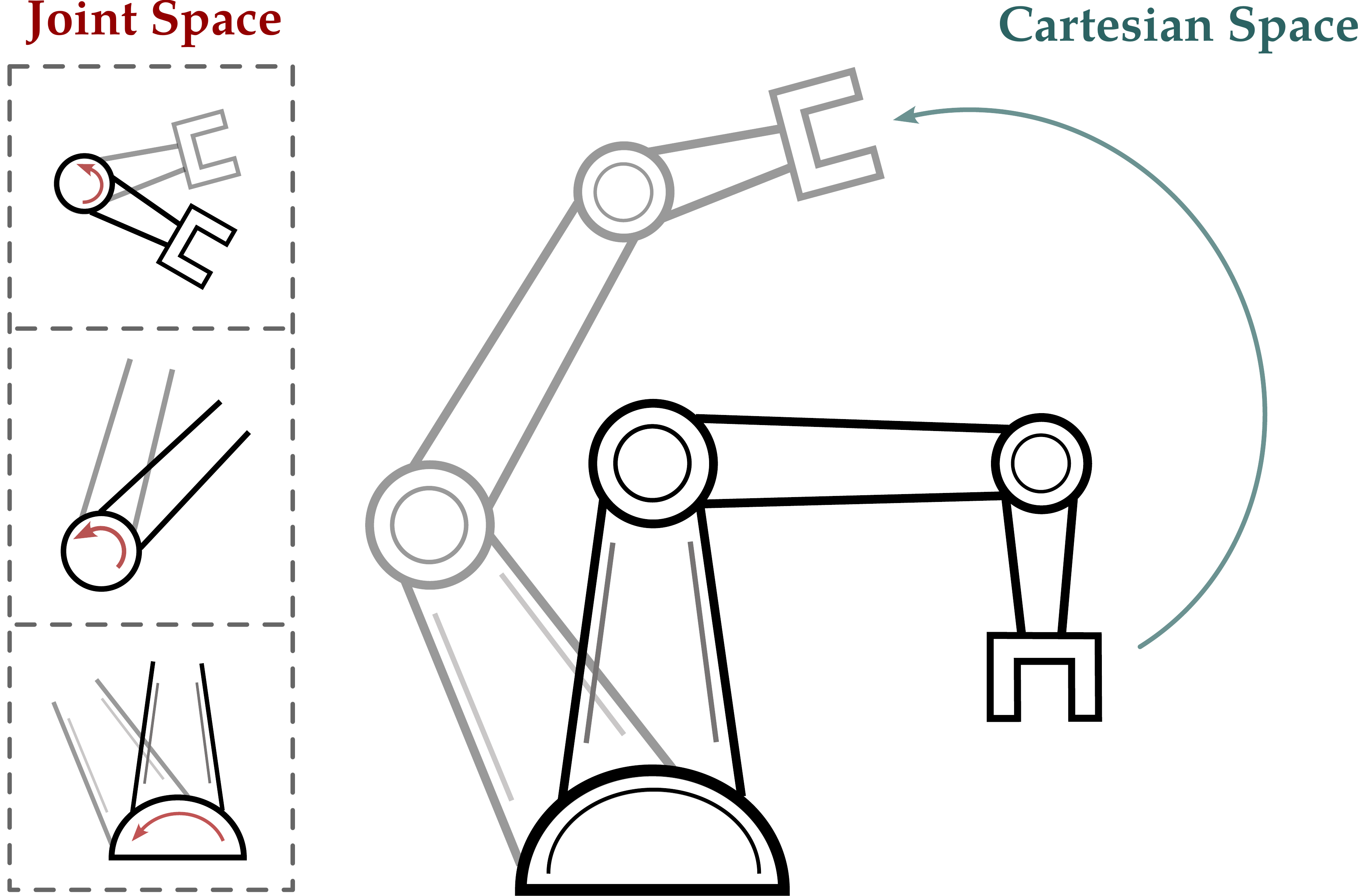}
\caption{Illustrtive comparison of joint space and Cartesian space.\label{fig:joint_cart}}
\end{figure}

Here we discuss the concept of learning spaces when representing demonstration data. The learning space not only encompasses where the training data from demonstrations are represented but also serves as the environment where the learning algorithm operates and generalizes the learned behavior. The choice of learning space is important as it provides background knowledge to the algorithm, thereby facilitating better learning and generalization within that designated space. While there are several possibilities for learning spaces, here we focus on two main choices commonly used for robotic manipulators (Figure~\ref{fig:joint_cart}).

\subsubsection{Joint Space}

The joint space of a robot is a comprehensive representation of its individual joint configurations. This space serves as the primary language of motor commands which offers a low-level and precise representation of the possible configuration of each joint \cite{ewerton2016incremental, barekatain2023dfl, pastor2009learning, tavassoli2023learning}.

\textbf{Learning in Joint Space:}
Learning in joint space offers several advantages for \ac{LfD} algorithms. Existing in Euclidean space makes the underlying math and data processing more efficient and straightforward. Additionally, since joint space directly corresponds to the robot's control layer, learned behaviors can be seamlessly integrated into the controller, which further simplifies the process.

However, a potential downside of learning in joint space is the risk of overfitting to specific demonstrations, which limits the robot's generalizability. Furthermore, focusing on joint space learning neglects to capture higher-level contextual information which relate to the semantics of the task. Finally, joint space learning is sensitive to hardware variations, making it difficult when it comes to scalability and transferring skills between different robots.

\textbf{Demonstrating in Joint Space:}
One notable advantage of demonstrating in joint space is the richness of the information obtained during the demonstration process, including the precise configuration of each individual joint. This level of detail is particularly advantageous for kinematically redundant manipulators, which can optimize null-space motion and obstacle avoidance.

However, a drawback lies in the intuitiveness of the learning process for human teachers. While joint space offers rich data for the robot to learn from, understanding how the learning algorithm learns and generalizes from this data is not intuitive for humans. The complexity involved in translating joint configurations into meaningful task representations can pose challenges for human teachers in understanding the learning process and predicting how the robot generalizes its learned skills.

In terms of acquiring demonstration data, kinesthetic teaching and teleoperation are straightforward methods for obtaining joint space data since joint states can be directly recorded. However, passive observation requires an additional step to convert raw information into joint space data, making it less practical to operate directly in joint space for this approach. In such cases, adding unnecessary transformations to obtain joint data is not justified, as it complicates the demonstration process.

\subsubsection{Cartesian Space}

This section introduces Cartesian space, a mathematical representation of three-dimensional physical space using orthogonal axes. In robotics, Cartesian space is commonly employed to describe the position and orientation of a robot's end-effector. \ac{LfD} in Cartesian space involves training robots to imitate demonstrated tasks or movements by utilizing the coordinates of the end-effector \cite{roveda2020assembly, meszaros2022learning, wu2023prim, johannsmeier2019framework, simonivc2021analysis, wu2021learning, lee2019making, maeda2017active, chisari2022correct, hu2022robot}.

\textbf{Learning in Cartesian Space:}
Cartesian space offers a natural representation for tasks involving end-effector movements and positioning, simplifying the learning process by directly addressing the space where tasks are being operated and executed. This choice is particularly advantageous for applications requiring precise end-effector control and potentially leads to better generalization in new situations. The consistency in representation allows for more effective generalization across different robotic systems and tasks.

One significant advantage of Cartesian space is the consistency of the dimension of the end-effector pose state across various robot platforms, regardless of their joint configurations. This standardized representation facilitates a more uniform approach to learning compared to joint space, which can vary in dimensionality from one robot to another. However, challenges arise when dealing with rotations within Cartesian coordinates, as rotation resides in a different manifold. It is required to utilize the calculus in a non-Euclidean space. This can increase the computational complexity of learning algorithms compared to those in the straightforward calculations of joint space.

Furthermore, the outcome of learning in Cartesian space cannot be directly integrated into the robot's controller. A transformation step is required to convert the learned information into joint space, adding an extra layer of complexity to ensure seamless integration into the robot's control system.

\textbf{Demonstrating in Cartesian Space:}
Cartesian space is an intuitive space for the human teacher, which facilitates a better understanding of how the learning process happens. Additionally, this intuitiveness enables easier inclusion of task parameters and better iterative feedback to the \ac{LfD} outcome, allowing the instructor to refine and optimize the robot's performance over successive teaching sessions.

However, a limitation arises in the encoding of end-effector behavior exclusively.
In redundant manipulators, the representation in Cartesian space does not directly consider null space motion.
The null space, which represents additional degrees of freedom beyond the end-effector behavior, is not explicitly encoded in Cartesian space.
This limitation restricts the teacher's ability to convey and refine complex motions involving redundant manipulators, potentially overlooking certain aspects of the robot's capabilities.

Finally, acquiring training data in Cartesian space via kinesthetic teaching requires having the kinematic model of the robot including the end effector hand or tool, and applying forward kinematic to the raw joint readings.
The approach is similar in teleoperation, although it depends on the design of the teleoperation interface.
Passive observation, however, demands the development of a mapping between the human hand or held tool and the robot's end effector.
Leveraging pattern recognition and feature extraction techniques, the movements of the human hands are interpreted and translated into end-effector movements, serving as valuable Cartesian-space training data.

\subsubsection{Remarks}
While joint space and Cartesian space are the most common and fundamental spaces to represent a task, there are a variety of choices that can be particularly designed and developed for the designated task.
For example, in  \cite{pignat2019bayesian}, the pixel space of the camera was used as a measure of distance between the peg and hole. 
End-to-end training on visual images was used in  \cite{johns2021coarse}.
Also, in  \cite{origanti2021automatic}, while the task is learned in Cartesian space, the Cartesian frame changes at each time step in a way that the z-axis always points towards the direction in which the force has to be applied.

When choosing approaches where a cost or reward function is involved, the design of such functions depends on the choice of latent/feature space, \textit{i.e.} the space that the reward or cost function has to represent.
In  \cite{biyik2023active}, the reward function is learned from demonstration, to represent as latent space of the training data. 
Later, the reward function is used by Reinforcement Learning to learn a suitable policy.
Likewise, in  \cite{yin2019ensemble}, a cost function is learned via Inverse Optimal Control, which represents the task's requirement for successful execution.

\subsection{Learning Methods}
Here we provide a comparative analysis of the most common learning methods used for \ac{LfD}. For each method, we discuss their learning concept and their training procedure, as well as their characteristics, strengths and weaknesses. Moreover, in Table \ref{tab:learning_comparison}, we present a summary of our comparative analysis. The table evaluates the learning methods across several key metrics to provide insight into their respective practical strengths and weaknesses in the manufacturing context. The metrics assessed include the implementation effort, explainability, generalization capability, training data efficiency, and safety and robustness. Implementation effort helps evaluate the practicality and resource requirements of integrating a particular learning method into manufacturing processes. Explainability assesses how well the reasoning behind the learned behaviors can be easily understood and interpreted by operators. Generalization capability indicates the extent to which a learning method can adapt to new or unseen situations, enhancing its versatility and applicability. The efficiency of training data usage highlights how effectively a method can learn from limited datasets, optimizing resource utilization and reducing data acquisition costs. Finally, Safety and robustness metrics assess the reliability and resilience of learned behaviors in the face of uncertainties.

\subsubsection{\ac{MP}}

\textbf{Learning Concept:}
A \ac{MP} encapsulates a low-level robot behavior defined by a trajectory generator and an exit condition.
The trajectory generator specifies the desired robot motion, while the exit condition determines when the movement should stop \cite{wu2023prim, johannsmeier2019framework, tavassoli2023learning, zhou2020movement}.
For instance, a typical \ac{MP} involves moving a robot until contact with a surface, where the trajectory generator guides the robot until a predefined force threshold is reached, signaling the exit condition. 
\acp{MP} do not require demonstration at the subtask level. They are manually designed and optimized by robotic experts. Instead, \ac{LfD} focuses on the task hierarchy, where the sequence in which to execute these pre-defined \acp{MP} is demonstrated to achieve a full task. Tasks composed of \acp{MP} can be structured using various methods such as state machines or task graphs. The essence of \acp{MP} is to offer a structured and optimized way to encode low-level robot behaviors that can be combined to achieve more complex tasks.

\textbf{Training Procedure:}
The design, implementation, and optimization of \acp{MP} is performed by the robotic expert, while the human teacher demonstrates the task hierarchy composed of \acp{MP}. In this way, subtasks are structured and tailored to specific tasks, resulting in efficient, reliable, and predictable behavior, while the demonstration effort is minimized. 
However, the manual design and tuning of \acp{MP} by robotic experts limit flexibility and control over lower-level behaviors for teachers. While \acp{MP} enhance learning performance by constraining the learning space to predefined combinations, their effectiveness depends on expert tuning, making them less adaptable to new behaviors, especially at the subtask level. \acp{MP} restricts generalization at the subtask level, demanding expert intervention to encode, tune, and integrate new behaviors into the \acp{MP}.

\subsubsection{\ac{DMP}}

\textbf{Learning Concept:}
The \ac{DMP} combines a spring damper dynamical system, known as an attractor model, with a nonlinear function to achieve a desired goal configuration \cite{ijspeert2013dynamical, saveriano2023dynamic}. The attractor model ensures convergence to the goal configuration, with its attractor point serving as the target goal. Without the nonlinear function, the model asymptotically converges to the goal. The role of the nonlinear function is to encode a certain behavior represented via the demonstration. By superposition of the nonlinear function and the attractor model, \ac{DMP} replicates the demonstrated behavior. Essentially, the nonlinear function guides the attractor system towards the desired behavior, while eventually vanishing as the system reaches the goal \cite{kastritsi2018progressive,nemec2013velocity, xing2023dynamic, simonivc2021analysis, si2022adaptive, origanti2021automatic, shaw2022constrained, sidiropoulos2023novel, sidiropoulos2021reversible, abu2022unified}.

\textbf{Training Procedure:}
To effectively train a \ac{DMP}, the training data should primarily exhibit temporal dependence, with time progression as the independent variable (x-axis) and the dependent variable (y-axis) representing aspects such as position, force, or stiffness. Subtask learning involves collecting training data by creating trajectories from joint readings, force sensors, or stiffness profiles. For whole task sequences, teaching data must first be segmented into subtasks, with each subtask fitted with its own \ac{DMP}. While \acp{DMP} are more suited for learning subtasks, approaches exist where multiple \acp{DMP} can be integrated into a single model to represent entire task sequences \cite{saveriano2019merging}. Motion-based and contact-based learning are feasible via \acp{DMP}, utilizing position trajectories \cite{sidiropoulos2023novel,simonivc2021analysis}, force trajectories \cite{han2022modified}, or stiffness profiles \cite{chang2022impedance,liao2022dynamic} as training data. Notably, \acp{DMP} offer the advantage of requiring only a single demonstration to generate a training set and train the model, although multiple demonstrations can be utilized for a more comprehensive training set \cite{han2022modified, ugur2020compliant}. Moreover, \acp{DMP} have been employed in context-dependent learning scenarios for collaborative tasks via points or task parameters \cite{nemec2018human,sidiropoulos2022dynamic}.

This training procedure involves representing time as a phase variable to make \ac{DMP} systems autonomous from direct time dependence. Training data, along with their derivatives, are input into the \ac{DMP} equation to generate target values for approximating the nonlinear term. Locally Weighted Regression (LWR) \cite{cleveland1988locally} is a typical choice for the nonlinear function approximator. The learned outcome is a nonlinear function mapping the phase variable to scalar values. The attractor model of \ac{DMP} is designed to ensure critical damping, facilitating convergence to the goal without oscillation. The core idea of \ac{DMP} lies in representing behavior as a deterministic system augmented by forcing terms, enabling learning at a higher level. While \acp{DMP} offer simplicity and reliability in implementation and generalization, their generalization mechanism is limited to a region around the original demonstration's configurations. Hyperparameters, although manually tuned, can be applied across various demonstrations without re-tuning. \acp{DMP} can be trained in joint space or Cartesian space, with multiple \acp{DMP} synchronized via the phase equation for joint space training, and a modified version for rotational values in Cartesian space represented using quaternions.

\begin{table}[H]
\caption{Comparison of learning methods in manufacturing contexts.\label{tab:learning_comparison}}
	\begin{adjustwidth}{-2cm}{0cm}
		\newcolumntype{C}{>{\centering\arraybackslash}X}
		\begin{tabularx}{\linewidth}{CCCCCCCCC}
            \toprule
            \textbf{Metric} & \textbf{MP} & \textbf{DMP} & \textbf{RL} & \textbf{GP} & \textbf{GMM} & \textbf{ProMP} \\
            \midrule
            Concept & Predefined deterministic behavior & Deterministic system with nonlinear forcing term & Interactive learning of reward and policy models & Probabilistic modeling of functions & Mixture of multiple Gaussians & Basis functions to model behavior \\
            \midrule
            Implementation Effort & Moderate & Low & High & Moderate & Moderate & Moderate \\
            \midrule
            Explainability & High & High & Low & High & High & Moderate \\
            \midrule
            Generalization Capability & Low & Moderate & High & Moderate to High & Moderate to High & Moderate to High \\
            \midrule
            Training Data Efficiency & High & High & Low & Moderate & Moderate & Low \\
            \midrule
            Safety and Robustness & Moderate to High & Low & Moderate to High & Moderate & Moderate & Moderate \\
			\bottomrule
		\end{tabularx}
	\end{adjustwidth}
\end{table}

\subsubsection{\ac{RL}}

\textbf{Learning Concept:}
\ac{RL} in the context of \ac{LfD} involves training robots to execute tasks by interacting with their environment, receiving feedback in the form of rewards or penalties, and adjusting their actions to maximize cumulative rewards \cite{peters2011towards, wu2021learning, lee2019making, vidakovic2020accelerating, biyik2023active, tsai2020constrained, chisari2022correct, wang2023learning, ma2020efficient}. At its core, \ac{RL} involves an agent (the robot) learning optimal actions to achieve a predefined objective within a given environment. This learning process hinges on the development of a policy, a strategy that guides the agent's actions based on the current state of the environment. The policy is refined through the optimization of a value function, which estimates the expected cumulative reward for specific actions in particular states. Rewards serve as feedback, reinforcing desirable actions and discouraging undesired behavior, thereby shaping the agent's learning trajectory. \ac{RL} algorithms balance exploration and exploitation, enabling the robot to discover effective strategies while leveraging known successful actions. However, \ac{RL} alone does not suffice for successful \ac{LfD} without an accurate reward function encapsulating the task requirements.

\ac{IRL} acts as a bridge between \ac{RL} and \ac{LfD}. Via human demonstrations, \ac{IRL} seeks the underlying implicit reward structure that guides those actions. The fundamental idea is to reverse engineer the decision-making process of the human teacher. Capturing the latent reward function enables the robot to replicate and generalize learned behavior to achieve similar goals in diverse contexts \cite{das2021model, alakuijala2023learning, trinh2024autonomous}.


\textbf{Training Procedure:}
To Train and \ac{LfD} algorithm via \ac{RL}-based methods, two key components are essential: a reward function and a policy function. The reward function encapsulates the task's definition, requirements, and success metrics, essentially encoding all the information provided by the teacher. Meanwhile, the policy function serves as the brain of the robot, dictating its behavior to execute the task. Training proceeds in two stages: first, designing or learning an appropriate reward function that accurately represents the desired task features and requirements, and second, training an \ac{RL} algorithm to learn a policy function based on this reward function through interaction with the environment.

During the first stage, the focus lies on devising a reward function that encapsulates the teacher's instructions regarding the task. While this function can be manually designed, a more comprehensive \ac{LfD} solution involves learning the reward function from human demonstrations \cite{wu2021learning, biyik2023active, vecerik2017leveraging, escontrela2024video}. The effectiveness of the \ac{LfD} solution is directly linked to how well the reward function encapsulates the task's key aspects. In the second stage, assuming a reward function is already established, an \ac{RL} algorithm learns a policy function by iteratively refining its behavior based on feedback from the environment and rewards obtained from the reward function.

Training via \ac{RL}-based approaches offers flexibility in encoding information and learning skills, with training data ranging from raw images to robot trajectories. However, it requires careful engineering of the reward function and task analysis by robotic experts. Additionally, \ac{RL}-based learning requires a dataset, not just a single demonstration, and involves modeling the environment, adding complexity to the implementation of \ac{LfD} algorithms in this manner. Tuning hyperparameters associated with the policy and reward functions, as well as potentially modifying the environment model for each task, are the steps that require robotic experts to ensure successful and efficient learning.

\subsubsection{\ac{GP}}

\textbf{Learning Concept:}
\acp{GP} are a probabilistic modeling approach in machine learning, capturing entire functions through mean and covariance functions known as kernel functions. The kernel function in \acp{GP} determines the similarity between function values at different points. This allows \acp{GP} to capture intricate patterns and relationships in data, while also estimating the uncertainty in those predictions. \acp{GP} are non-parametric, which means they are capable of learning from limited data points while being able to adjust their complexity with more data. Predictions from \acp{GP} include both anticipated function values and associated uncertainty, particularly useful in scenarios with sparse or noisy data \cite{meszaros2022learning, maeda2017active, arduengo2023gaussian, jaquier2020learning}.

The conceptual advantage of learning methods like \ac{GP} is their ability to quantify uncertainty, which is important for understanding the model's confidence in its predictions. This feature enhances human comprehension of the learning process and can serve as a safety mechanism for robots, where uncertain policies could lead to errors or damages. By monitoring the model's uncertainty and providing feedback, human teachers can refine the \ac{GP}'s behavior and adjust uncertainty bounds accordingly, ensuring safer and more robust execution.


\textbf{Training Procedure:}
The training process for \ac{GP} models involves learning from input-output pairs, where the independent variable can be any sequential variable, such as time or the position of the end effector. However, it is important to maintain the sequential nature of the data, \textit{e.g.}, restricting scenarios where the end effector revisits a location. \acp{GP} excel at learning subtasks with limited demonstration data, even one-shot demonstration input. To improve the generalization capability, it is advisable to train \ac{GP} in Cartesian space. This is because small changes in joint values can result in significant changes in the end effector in joint space. Moreover, training in joint space makes the uncertainty measures less interpretable.

\ac{GP} is not particularly demanding in terms of implementation and hyperparameter tuning. The robot expert needs to design a kernel function in the formulation of \ac{GP}, as well as very few hyperparameters. Moreover, \ac{GP} is flexible across various tasks, \textit{i.e.} it does not often require significant hyperparameter tuning or design alterations when transitioning from one task to another.

\subsubsection{\ac{GMM}}

\textbf{Learning Concept:}
\acp{GMM} offer a probabilistic method similar to GPs for modeling functions, representing them as a mixture of multiple Gaussian distributions. Each Gaussian component within a \ac{GMM} represents a cluster in the dataset. In the context of \ac{LfD}, \acp{GMM} can be used to model the underlying structure of human demonstrations. Due to their multi-modal nature, \acp{GMM} can capture complex behaviors while being flexible in terms of the number of learning variables \cite{pignat2019bayesian, biyik2023active, zhu2022learning}. They can also encode variability in demonstrations, giving a measure of accuracy at each point. similar to GPs. Additionally, \acp{GMM} can be updated locally with new data without affecting other segments of the learned behavior. This allows \acp{GMM} to naturally cluster data and learn complex behaviors from human demonstrations.

\textbf{Training Procedure:}
\acp{GMM} allow for flexible training variable dimensions through multivariate Gaussian distributions, enabling each distribution to represent and train on different aspects of a task or subtask. Unlike GPs which can learn from single demonstration, \acp{GMM} require multiple demonstrations to capture the statistics and effectively learn the task. Additionally, \acp{GMM} are typically more intuitive when learned in Cartesian space rather than joint space, similar to GPs.

\subsubsection{\ac{ProMP}}

\textbf{Learning Concept:}
\ac{ProMP} is a learning approach developed for \ac{LfD} which employs Gaussian basis functions to model demonstrated behavior \cite{paraschos2013probabilistic}. The parameters of \ac{ProMP} are typically weights associated with the basis functions. \ac{ProMP} introduces a probabilistic aspect into learning, similar to GP and GMM, allowing the model to accommodate uncertainty in learned movements and generalize them to different conditions or contexts \cite{ewerton2016incremental, koert2019learning, kulak2021active, yue2024probabilistic}.

\textbf{Training Procedure:}
The training procedure for \acp{ProMP} involves representing training data as time-series trajectories, with each trajectory corresponding to a specific demonstration of the task. These trajectories are often normalized or preprocessed to ensure consistency across different demonstrations. Through an optimization process, the model seeks to find the parameters that best fit the observed trajectories from demonstrations, constructing a probabilistic model capturing the distribution of trajectories and their likelihood at each point in time. However, \acp{ProMP} generally require a larger training dataset compared to other probability-based \ac{LfD} methods.

\subsubsection{Remarks}
In addition to the mentioned methods, various other learning approaches serve as the core of \ac{LfD} algorithm, with customization based on task-specific requirements. For example, \acp{HMM} are commonly utilized for learning behaviors, as used in  \cite{roveda2020assembly}. Additionally, methods based on optimal control are also employed, which are conceptually similar to \ac{RL}. For example, in  \cite{yin2019ensemble}, \ac{IOC} is used to learn the cost function of the task, similar to \ac{IRL}. This function is later used to find an optimal policy to generalize the desired task.

\section{How to Refine}
\label{sec:HowToRefine}

The final step after completing the design and development of an \ac{LfD} process, is to analyze and evaluate their performance, which guides the question of ``How to Refine". This section dives into the main key trends and directions within the state-of-the-art that aim to refine \ac{LfD} algorithms across various aspects. For each trend, we will explore how it can improve \ac{LfD} performance and identify potential areas for further research. The goal is to provide insights into possible research objectives for evaluation and improvement. This analysis will provide a refined perspective on the previously discussed aspects, ultimately contributing to an iterative loop of improvement for the entire \ac{LfD} process.

\subsection{Learning and Generalization Performance}

Although current \ac{LfD} approaches can learn from human teachings and generalize the behavior to new situations, it is still far from the idea of learning behavior that can be seen from humans. If human learning capabilities are considered the ideal case, \ac{LfD} approaches are not even close to cognitive learning capabilities. Therefore, it is a crucial line of refinement on the \ac{LfD} approaches to get them closer to the cognitive learning power of human beings.
The performance of learning and generalization refers to how well the algorithm can capture the essence of the desired behavior from human demonstration, and how intelligently the algorithm generates essentially the same behavior but adapts to the new environment or situation or context condition. Learning and generalization are two entangled factors that directly affect each other. A better the learning performance subsequently leads to a better generalization, and improving generalization essentially means that learning performance has been improved.
While using typical state-of-the-art \ac{LfD} approaches already performs well in learning and generalization, they operate under assumptions, and actually cannot generalize to every possible case or situation. That is why it is necessary to improve the \ac{LfD} approach based on what we require for our task and what is missing in the current \ac{LfD} approaches. 
Improving learning performance means improving how the human demonstrations are processed by the learning algorithm, as well as modifying the core algorithm of learning to better capture different aspects of the behavior from the demonstrations. One trend is the approach of incremental learning  \cite{maeda2017active, auddy2023continual, ewerton2016incremental, simonivc2021analysis, johns2021coarse}. Incremental learning refers to the ability of a robot to continuously acquire and refine its knowledge and skills over time as it interacts with its environment or receives additional demonstrations from a human operator. In  \cite{maeda2017active}, A \ac{GP} is learned via one demonstration, but more demonstrations are provided through the operation of the robot to further refine \ac{GP} training and improve its learned behavior. As another example, authors in  \cite{auddy2023continual} focused on continual learning for teaching letters in the alphabet incrementally without the algorithm forgetting the previously learned letters. The algorithm was able to write ``Hello World" at the end, with the accumulated knowledge.

Building upon incremental learning, the concept of interactive learning emerges  \cite{celemin2019interactive, ewerton2016incremental, kulak2021active, si2022adaptive}. Interactive learning refers to a learning paradigm in which the robot actively engages with the human teacher or the learning environment to acquire knowledge or skills. Unlike passive learning methods where information is simply presented to the learner, interactive learning involves two-way communication and dynamic interaction between the learner and the learning material. In  \cite{celemin2019interactive}, authors provide an interactive learning framework through which non-expert human teachers can advise the learner in their state-action domain. In  \cite{ewerton2016incremental}, the human teacher kinesthetically corrects the robot's trajectory during execution to teach the robot to perform the task accurately. A joint probability distribution of the trajectories and the task context is built from interactive human corrections. This distribution is updated over time, and it is used to generalize to the best possible trajectory given a new context. 

Another technique is Active Querying  \cite{kulak2021active,maeda2017active, biyik2023active}. In this technique, the learner dynamically decides which data points or demonstrations are most informative for the learning or decision-making process and requests the corresponding information from the teacher. This approach is particularly helpful for improving performance in an efficient way and acquiring the most relevant information. In  \cite{kulak2021active}, they focused on the demonstration distribution when training \acp{ProMP}, and the fact that it is not trivial how to add a good demonstration in terms of improving generalization capabilities. So they learn a \ac{GMM} over the demonstrations distribution. and use epistemic uncertainty to quantify where a new demonstration query is required. Their proposed active learning method iteratively improves its generalization capabilities by querying useful and good demonstrations to maximize the information gain. In a similar way, the uncertainty measure of \acp{GP} is used in  \cite{maeda2017active} to trigger a new demonstration request.

Aside from the mentioned learning paradigms, it is often required to modify learning algorithms based on the application use case or the context in which the \ac{LfD} algorithm is employed. For example, there are several works to improve the performance of \ac{LfD} with respect to contact-rich tasks  \cite{wu2023prim, johannsmeier2019framework,si2022adaptive, shi2021combining}. In  \cite{johannsmeier2019framework}, they proposed an approach to reduce the learning time of insertion tasks with tolerances of up to sub-millimeters, with application in manufacturing use cases. In  \cite{wu2023prim}, a novel task similarity metric is introduced, and it is used to generalize the already-learned insertion skills to novel insertion tasks without depending on domain expertise. In another context,  \cite{vuong2021learning} considers the assembly use cases and explores the idea that skillful assembly is best represented as dynamic sequences of manipulation primitives, and that such sequences can be automatically discovered by Reinforcement Learning. The authors in  \cite{cui2022coupled} extended \acp{DMP} to manipulating deformable objects such as ropes or thin films, to account for the uncertainty and variability from the model parameters of the deformable object. Another important aspect of learning is to learn factor which are in non-euclidean spaces. In  \cite{abu2022unified}, the formulation of \ac{DMP} is modified to become geometry aware, so that the new formulation can be adapted to the geometric constraints of Reimanian manifold. 

Finally, for certain contexts, some works attempt to design and develop control schemes to be learned in order to better learn and execute the behavior. The work in  \cite{origanti2021automatic} has focused on the fact that some tasks such as grinding, sanding, polishing, or wiping require a hybrid control strategy in order to accurately follow the motion and the force profile required for successful task execution. To enhance the learning process of these tasks, they proposed some pre-programmed control strategies with parameters to tune via non-expert demonstrations. Such parameters are extracted from one-shot demonstrations and learn the motion-force task more efficiently.

Aside from the context, several works have attempted to provide improvements on the learning performance of a certain algorithm. Such algorithm-based improvements are dedicated to resolve the performance issues inherent in a learning method. In  \cite{sidiropoulos2021reversible}, They proposed a new formulation for \acp{DMP} in order to make it reversible in both directions and make it more generalizable. Authors in  \cite{simonivc2021analysis} introduce Constant Speed Cartesian \acp{DMP}, which completely decouples the spatial and temporal components of the task, contributing to a more efficient learning. In their \ac{LfD} approach based on \ac{IOC}, the work in  \cite{yin2019ensemble} proposed an ensemble method which allow for more rapid learning of a powerful model by aggregating several simpler \ac{IOC} models. The work in  \cite{maeda2017active} combined \acp{GP} with \acp{DMP}, as \acp{GP} do not guarantee convergance to an arbitrary goal. Therefore, a \ac{DMP} is learned on the \ac{GP} output to ensure that any arbitrary goal is reached. With respect to Deep learning based approaches,  \cite{lee2019making} has focused on reducing the sample complexity by introducing self-supervised methods. For \ac{RL}-based approaches, authors in  \cite{vecerik2017leveraging} have developed a method to use demonstrations for improving learning sparse-reward problems. Both demonstrations and interactions are used to enhance the performance of learning.

\subsection{Accuracy}

While accuracy and precision can be mainly improved by improving the learning performance, it is not necessarily sufficient to achieve the desired accuracy by focusing generally only on the learning performance. While improving learning can enhance the overall generalization performance of the algorithm in the case of new scenarios, it is not a guarantee that the generalization outcome is accurate in terms of the desired task's metrics. Not only does the teaching process influence the accuracy of the outcome, but the execution strategy of the \ac{LfD} is also the final stage that plays a major role in the outcome.

The notion of accuracy has a subjective nature, \textit{i.e.} it can be defined differently from task to task. Therefore, there is no unified definition to describe the metric of accuracy across arbitrary tasks. However, several factors can be associated generally with the metrics of accuracy, to measure how accurate is the \ac{LfD} algorithm, to some extent, with respect to the outcome. One of these factors is the success rate. Over various execution of the task via \ac{LfD} policy, the success rate of the task to achieve a desired goal, can be a simple yet effective measure of how accurately a task is executed. If the accuracy requirements are not satisfied until a threshold, the task will not succeed at the end. Hence, success rate can be a measurable factor to describe the accuracy of a task, and a tangible metric to work towards improving by enhancing accuracy.

Besides focusing on the learning algorithm's performance, there are more dedicated approaches and trends to improve the accuracy of an \ac{LfD} algorithm. One direction is to focus on the question of how to modify the teaching and demonstration method to enable human teachers to teach the task more accurately  \cite{perico2019combining, nemec2018human, nemec2018efficient, ewerton2016incremental, meszaros2022learning}. For example, In  \cite{nemec2018human}, they separated the demonstration of the shape of the trajectory from the timing of the trajectory. While it is cognitively demanding to demonstrate a motion with high accuracy and a high velocity at the same time, the ability to independently demonstrate the path enables the teachers to solely focus on teaching and refining the robot path and define the behavior in a more precise way. Moreover, in  \cite{perico2019combining}, they have combined incremental learning with variable stiffness of the robot during kinesthetic feedback. They used the variability of the already-captured demonstrations to adjust the stiffness of the robot. When there is low variation in a region, \textit{i.e.} higher accuracy, the robot is more stiff, allowing the teacher to provide smaller adjustments, while regions with higher variability have lower stiffness, allowing the teacher to move the robot more freely.

Another approach is to focus on how the \ac{LfD} output plan is executed finally with the robot. Here, the focus is on execution strategies with the goal of improving the success rate of a task  \cite{hu2022robot, vidakovic2020accelerating, johannsmeier2019framework}. In  \cite{hu2022robot} they considered a small-parts assembly scenario, where the tolerances are comparatively low, which leads to more sensitivity to errors in misalignments due to tight tolerances. They have proposed an impedance control strategy to drive the robot along the assembly trajectory generated by \ac{LfD}, but also record and track a required wrench profile so that tolerance misalignments can be resolved with the compliance of the contact, and in this way increasing the success rate of the task and avoiding failures when there is contact due to tolerance errors. In  \cite{vidakovic2020accelerating}, the authors claim that \ac{LfD} output performs well in generalization but fails to maintain the required accuracy for a new context. Hence they use the \ac{LfD} output as an initial guess for an \ac{RL} algorithm designed according to the task's model, and refine the \ac{LfD} output find the optimal policy for the specific task.

\subsection{Robustness and Safety}

The general \ac{LfD} process is mainly concerned with successfully learning and executing a desired task with a specific accuracy, while it is crucial to consider how the process lifecycle is aligned with safety requirements and how well the system can handle unexpected scenarios. This is of extra importance in manufacturing cases where the robot shares an industrial environment alongside humans, with more potential dangers. Although \ac{LfD} can generalize to new scenarios, there is no guarantee that the devised task policy is aligned with safety requirements or passes through the safety region. It is also not guaranteed that in case of unforeseen situations during the task execution, the robot makes a safe decision to accommodate the new situation. Therefore, it is important to equip the \ac{LfD} processes with proper mechanisms to ensure robustness and safety in the robot's operational environment.

The concept of robustness gains meaning when there is a possibility of a fault, disturbance, or error throughout the process, that might mislead the learning process or cause the system to end up in an unknown state. In this case, a robust \ac{LfD} system is capable of reliably recovering from the imposed state and successfully finishing the task whatsoever. Alongside robustness, the concept of safety ensures that the robot's operations and decisions do not cause any harm to the humans working along, especially during the teachings and interactions. It also ensures that the operations remain in a safe region to prevent damage to the working environment, work objects, and the robot itself. It is evident that robustness and safety are essential components of \ac{LfD} systems that must be carefully addressed to enable their effective deployment in real-world applications.

One main aspect of robustness and safety in \ac{LfD} processes is related to the \ac{HRI}. Since \ac{LfD} lifecycle is mainly concerned with interaction with humans, it is evident that focusing on \ac{HRI} robustness and safety becomes a major trend on enhancing the reliability of \ac{LfD} systems  \cite{Yang2024enhancing}.

One main trend of improving safety and robustness is focused on \ac{HRI}  \cite{tsai2020constrained,chisari2022correct,meszaros2022learning,khoramshahi2019dynamical,nemec2018human}. This includes the teaching and demonstration as well as any other form of interaction throughout the process. With respect to robustness, in  \cite{tsai2020constrained}, to efficiently learn from suboptimal demonstrations, the paper proposes an \ac{RL}-based optimization where the demonstrations serve as the constraints of the optimization framework. the objective was to outperform the suboptimal demonstrations and find the optimal trajectory in terms of length and smoothness. The work in  \cite{chisari2022correct} proposes an interactive learning approach where instead of depending on perfect human demonstrations to proceed with learning, the human can interactively and incrementally provide evaluative as well as corrective feedback to enhance the robustness of learning against imperfect demonstrations. In terms of safety, in  \cite{meszaros2022learning} kinesthetic teaching is replaced with teleoperation to enhance safety while providing local corrections to the robot. This is because kinesthetic teaching can become more dangerous with increasing the robot's velocity of execution. Moreover, in many works such as  \cite{khoramshahi2019dynamical,nemec2018human} the control scheme is based on impedance control to guarantee compliant interactions with humans, avoid sudden unexpected motions, and improve \ac{HRI} safety.

Another trend for improving robustness is focused on the robustness against various disturbances and errors through out the \ac{LfD} cycle  \cite{eiband2023collaborative,pignat2019bayesian, lee2019making}. The work in  \cite{eiband2023collaborative} has focused on the fact that while \ac{LfD} can allow non-experts to teach new tasks in industrial manufacturing settings, exerts are still required to program fault recovery behaviors. They have proposed a framework where robots autonomously detect an anomaly in execution and learn how to tackle that anomaly by collaborating with human. They represent a task via task graphs, where a task is executed from start to end. If the robot detects an anomaly, it waits and asks humans whether to demonstrate a recovery behavior or refine the current execution behavior. In case of learning a new recovery behavior from the teacher, a conditional state is added to the graph at the point of anomaly, and next time the robot checks for the condition to see if it should continue the normal execution or switch to the recovery behavior. Similarly, the authors in  \cite{pignat2019bayesian} proposed Bayesian \ac{GMM} to quantify the uncertainty of the imitated policy at each state. They fuse the learned imitation policy with various conservative policies in order to make the final policy robust to perturbations, errors and unseen states. For example, in a board wiping task, the imitation policy learned the force profile required to wipe the board, while the conservative policy ensured circular motion on the board. together, they make the board wiping policy robust so that the motion is desirable while applied force is enough to actually wipe the board.

Lastly, there are several works focusing on robustness and safety considering the limitations in the robot's operating environment  \cite{shaw2022constrained, sidiropoulos2023novel}. In  \cite{shaw2022constrained}, the problem of collision avoidance was considered. They proposed a modified formulation of \ac{DMP}, where they incorporated a zeroing barrier function in the formulation and solved a nonconvex optimization in order to find a collision free path through their constrained \ac{DMP}. From another perspective, the work in  \cite{sidiropoulos2023novel} addressed the issue that there is no guarantee that \ac{LfD} outcome respect kinematic constraints when generalizing. so they formed a QP optimization problem that enforces the kinematic constraints and finds the \ac{DMP} weights where the optimal trajectory is closest to the original \ac{DMP} trajectory while respecting the constraints.
\section{Conclusion}
\label{sec:Conclusion}

This paper presented a structured approach to integrating \ac{LfD} into the roboticization process using manipulators for manufacturing tasks. It addressed key questions of "What to Demonstrate," "How to Demonstrate," "How to Learn," and "How to Refine," providing practitioners with a clear roadmap to implement \ac{LfD}-based robot manipulation. First, we identified the scope of demonstration based on the desired task's characteristics and determined the knowledge and skill required to be demonstrated to the robot. Then, based on the scope of demonstration, we explored demonstration methods and how human teachers can provide demonstrations for the robot, providing insights to extract the demonstration mechanism. Next, we focused on the learning approaches to enable efficient task learning and execution from the provided demonstrations. We first explored the possible learning spaces, followed by the common learning methods along with their pros and cons. Finally, we provided trends and insights to evaluate and improve the \ac{LfD} process from a practical point of view, giving research directions and objectives for building upon the state of the art.

By providing a detailed and structured analysis into determining the scope of demonstration, devising demonstration mechanisms, implementing learning algorithms, and refining \ac{LfD} processes, our review enables both researchers and industry professionals to develop application-based \ac{LfD} solutions tailored for manufacturing tasks. This paper offered a practical and structured guide, making \ac{LfD} accessible to practitioners with moderate expertise requirements. Through comprehensive questionnaire-style guidance, we provided step-by-step instructions and main research directions for refining \ac{LfD} performance in manufacturing settings, thus bridging the gap between research and practice in the field of robotic automation.


\section*{Acknowledgments}
This research was funded in whole by the Luxembourg National Research Fund (FNR), grant reference 15882013. For the purpose of open access, and in fulfilment of the obligations arising from the grant agreement, the author has applied a Creative Commons Attribution 4.0 International (CC BY 4.0) license to any Author Accepted Manuscript version arising from this submission.

\bibliographystyle{unsrt}  
\bibliography{references}

\end{document}